\documentclass{article}

\PassOptionsToPackage{numbers, compress}{natbib}

\usepackage[preprint]{neurips_2026} 

\usepackage[utf8]{inputenc} 
\usepackage[T1]{fontenc}    
\usepackage{hyperref}       
\hypersetup{
    hidelinks 
}
\usepackage{url}            
\usepackage{booktabs}       
\usepackage{amsfonts}       
\usepackage{amsmath}        
\usepackage{nicefrac}       
\usepackage{microtype}      
\usepackage[table]{xcolor}  
\usepackage{graphicx}       
\usepackage{subcaption}     
\usepackage{multirow}       
\usepackage{enumitem}       
\usepackage{makecell}       
\usepackage{wrapfig}        
\usepackage{tcolorbox}      
\tcbuselibrary{breakable}   
\usepackage[capitalise,noabbrev]{cleveref} 
\usepackage{orcidlink}

\graphicspath{{figures/}}

\newcommand{\gain}[1]{\textcolor{green}{#1}}

\usepackage{pifont}
\newcommand{\cmark}{\ding{51}}%
\newcommand{\xmark}{\ding{55}}%

\usepackage{xspace}
\newcommand{\benchname}{\textsc{OmniPro}\xspace}

\title{OmniPro: A Comprehensive Benchmark for Omni-Proactive Streaming Video Understanding}

\author{
\textbf{Ruixiang Zhao$^{1}$\orcidlink{0009-0008-9984-1841} \quad Jie Yang$^{2,}$\thanks{Corresponding authors: Xirong Li (xirong@ruc.edu.cn), Jie Yang (cvjieyang@tencent.com)} \quad Zijie Xin$^{1}$\orcidlink{0000-0002-9220-8735} \quad Tianyi Wang$^{2}$} \\
\textbf{Fengyun Rao$^{2}$ \quad Jing LYU$^{2}$ \quad Xirong Li$^{1,*}$\orcidlink{0000-0002-0220-8310}} \\[1mm]
$^{1}$Renmin University of China \hspace{6mm} $^{2}$WeChat Vision, Tencent Inc. \\[2mm]
\textbf{Project page}: \textcolor{purple}{\href{https://ruixiangzhao.github.io/OmniPro/}{https://ruixiangzhao.github.io/OmniPro}}
}

\begin{document}

\maketitle

%

\begin{abstract}

Omni-proactive streaming video understanding, i.e., autonomously deciding when to speak and what to say from continuous audio-visual streams, is an emerging capability of omni-modal large language models. Existing benchmarks fall short in three key aspects: they rely primarily on visual signals, adopt polling or fixed-timestamp protocols instead of true proactive evaluation, and cover only a limited range of tasks, preventing reliable assessment and differentiation of omni-proactive streaming models.
We present \benchname, the first benchmark to jointly evaluate omni-modal perception, proactive responding, and diverse video understanding tasks. It comprises 2,700 human-verified samples spanning 9 sub-tasks and 3 cognitive levels, covering 6 basic video understanding capabilities. Notably, 84\% of samples require audio signals (speech or non-speech), and each sample is annotated with modality-isolation labels to enable fine-grained multimodal analysis. 
We further introduce a dual-mode evaluation protocol: \textit{Probe} mode assesses content understanding by querying the model before and after each ground-truth trigger, while \textit{Online} mode evaluates full proactive ability by requiring models to autonomously decide when to respond in streaming input. 
Evaluating 11 representative models reveals three key findings: (1) audio provides consistent gains but with highly variable utilization across models, (2) performance degrades significantly over time, indicating limited long-horizon robustness, and (3) non-speech audio perception remains the weakest dimension.
\end{abstract}

\section{Introduction}
\label{sec:intro}

Omni-proactive streaming video understanding, i.e., autonomously deciding when to speak and what to say based on continuous audio-visual signals, is emerging as a core capability of omni multimodal large language models. Despite growing interest in streaming and multimodal modeling~\cite{videollm-online,streamo,mmduet2,streambridge,dispider,minicpm-o,livestar}, a fundamental question remains unanswered: \textit{\textbf{what constitutes a good omni-proactive streaming model}}? We argue that such a model must satisfy three key criteria: (1)~\textbf{Omni-modal perception}: it should jointly reason over visual signals, speech, and non-speech audio (e.g., environmental sounds), as real-world triggers are inherently multimodal. (2)~\textbf{Proactive responding}: it must decide when to respond without external polling or fixed schedules, which distinguishes proactive behavior from passive response. (3)~\textbf{Diverse video understanding tasks}: it should support a broad range of tasks beyond simple event alerting, including monitoring, grounding, counting, narration, and predictive reasoning, reflecting the complexity of real-world scenarios.

To assess these three criteria, a benchmark must be explicitly designed to test them in a unified framework. However, as shown in the left (blue-shaded) columns of \cref{tab:benchmark_comparison}, existing proactive streaming benchmarks\footnote{The ``-Pro'' suffix denotes the proactive evaluation subset of each original benchmark.} fall short across all three dimensions. For \textbf{omni-modal perception}, StreamingBench-Pro~\cite{streamingbench} and OVO-Bench-Pro~\cite{ovobench} rely exclusively on visual cues, while OmniMMI-Pro~\cite{omnimmi} involves only $\sim$35\% speech content with no non-speech sound; none can differentiate omni-modal models from vision-only counterparts. For \textbf{proactive responding}, StreamingBench-Pro polls the model every second and OVO-Bench-Pro queries the model at several preset time points; both remain essentially offline and do not allow the model to initiate responses on its own. Only OmniMMI-Pro lets the model freely decide when to respond, yet it permits only a single response per question, leaving multi-trigger decision-making untested. For \textbf{diverse video understanding tasks}, all three benchmarks exhibit severely limited coverage, capturing only a small fraction of the basic capability space. Overall, no existing benchmark simultaneously evaluates all three criteria, resulting in a clear evaluation gap that contrasts sharply with the rapid emergence of proactive streaming models.

\begin{table}[t]
\centering
\caption{\textbf{Benchmarks for proactive streaming video understanding}. Blue-shaded columns: evaluation capability along the three proposed criteria. Orange-shaded columns: dataset statistics. ``Resp./Ques.'': average responses per question. ``1st Resp.'': average first response time.}
\label{tab:benchmark_comparison}
\setlength{\tabcolsep}{4pt} 
\renewcommand{\arraystretch}{1.5} 
\resizebox{\linewidth}{!}{%
\begin{tabular}{@{}l ccc|rrrrrcc@{}}
\toprule
\multirow{2}{*}{\textbf{Benchmark}} & \multicolumn{3}{c}{\cellcolor{blue!8}\textbf{Evaluation Capability}} & \multicolumn{7}{c}{\cellcolor{orange!8}\textbf{Dataset Statistics}} \\
\cmidrule(lr){2-4} \cmidrule(l){5-11}
& \cellcolor{blue!8}Omni & \cellcolor{blue!8}Proactive & \cellcolor{blue!8}Diversity & \cellcolor{orange!8}\# Videos & \cellcolor{orange!8}Dur. (s) & \cellcolor{orange!8}\# Ques. & \cellcolor{orange!8}Resp./Ques. & \cellcolor{orange!8}1st Resp. (s) & \cellcolor{orange!8}Sound & \cellcolor{orange!8}Speech \\
\midrule
StreamingBench-Pro~\cite{streamingbench} & \cellcolor{blue!8}\xmark & \cellcolor{blue!8}\xmark & \cellcolor{blue!8}1/6 & \cellcolor{orange!8}50 & \cellcolor{orange!8}636 & \cellcolor{orange!8}250 & \cellcolor{orange!8}1.0 & \cellcolor{orange!8}9.5 & \cellcolor{orange!8}\xmark & \cellcolor{orange!8}\xmark \\
OVO-Bench-Pro~\cite{ovobench} & \cellcolor{blue!8}\xmark & \cellcolor{blue!8}\xmark & \cellcolor{blue!8}2/6 & \cellcolor{orange!8}134 & \cellcolor{orange!8}625 & \cellcolor{orange!8}172 & \cellcolor{orange!8}9.1 & \cellcolor{orange!8}29.2 & \cellcolor{orange!8}\xmark & \cellcolor{orange!8}\xmark \\
OmniMMI-Pro~\cite{omnimmi} & \cellcolor{blue!8}\xmark & \cellcolor{blue!8}\cmark & \cellcolor{blue!8}1/6 & \cellcolor{orange!8}400 & \cellcolor{orange!8}350 & \cellcolor{orange!8}400 & \cellcolor{orange!8}1.0 & \cellcolor{orange!8}36.4 & \cellcolor{orange!8}\xmark & \cellcolor{orange!8}\cmark \\
\midrule
\benchname & \cellcolor{blue!8}\cmark & \cellcolor{blue!8}\cmark & \cellcolor{blue!8}6/6 & \cellcolor{orange!8}1,262 & \cellcolor{orange!8}189 & \cellcolor{orange!8}2,700 & \cellcolor{orange!8}3.4 & \cellcolor{orange!8}54.1 & \cellcolor{orange!8}\cmark & \cellcolor{orange!8}\cmark \\
\bottomrule
\end{tabular}%
}
\end{table}

\begin{figure*}[t]
  \centering
  \includegraphics[width=\linewidth]{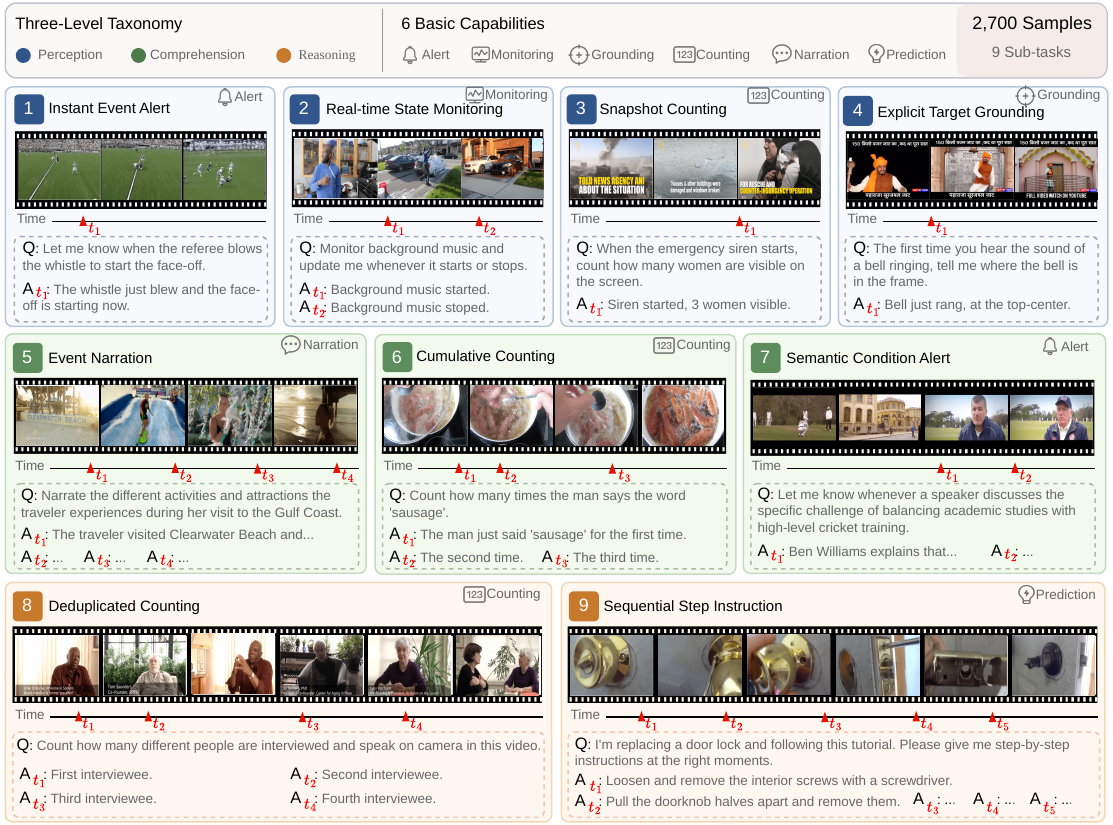}
  \caption{\textbf{Overview of \benchname}. The benchmark comprises 9 sub-tasks organized into three cognitive levels, collectively covering 6 basic video understanding capabilities. Each panel shows a representative sample with its video frames, time-aligned triggers (marked by red triangles), user instruction (\textbf{Q}), and expected proactive responses (\textbf{A}). Audio-dependent triggers are prevalent across tasks, requiring models to perceive both visual and auditory signals.}
  \label{fig:benchmark}
\end{figure*}

To address these limitations, we present \benchname, the first comprehensive benchmark for omni-proactive streaming video understanding. As illustrated in \cref{fig:benchmark}, \benchname contains 2,700 human-verified samples spanning 9 sub-tasks, organized into three cognitive levels that map to 6 basic video understanding capabilities. At the data level, 84\% of samples depend on audio information (speech or non-speech sound), and each sample carries modality-isolation labels enabling fine-grained multi-modal ablation. At the evaluation level, we introduce a dual-mode protocol: \textit{Probe} evaluates content understanding by querying the model before and after each ground-truth trigger time without requiring streaming capability, while \textit{Online} mode evaluates full proactive ability by requiring models to autonomously decide when to respond in a continuous video stream. Overall, \benchname is the first benchmark to jointly evaluate omni-modal perception, proactive responding, and diverse video understanding tasks within a unified framework.

We evaluate 11 representative models on \benchname, spanning open-source and proprietary systems in both probe and online modes. Key findings include: (1)~current omni models benefit from audio yet differ markedly in their utilization ability, with audio-visual input outperforming video-only input by +2.4 to +11.1 across models. (2)~performance degrades substantially as triggers occur later in the video, with models retaining on average only 37\% of their early-segment performance, indicating challenges in modeling long-term temporal dependencies. (3)~non-speech sound perception (e.g., environmental sounds) remains the weakest dimension across all models. These results demonstrate the discriminative power of \benchname and identify concrete open challenges for future research.

\noindent Our contributions are summarized as follows:
\begin{itemize}[leftmargin=*,nosep]
  \item \textbf{Benchmark}. We introduce \benchname, the first comprehensive benchmark for omni-proactive streaming video understanding, comprising 2,700 human-reviewed samples across 9 sub-tasks with 84\% audio dependency.
  \item \textbf{Taxonomy}. We design a hierarchical taxonomy across three cognitive levels that covers six basic video understanding capabilities. This framework enables a structured evaluation of omni-proactive streaming video understanding.
  \item \textbf{Evaluation}. We propose a dual-mode evaluation protocol: Probe for content understanding assessment and Online for full proactive ability evaluation.
  \item \textbf{Analysis}. We evaluate 11 representative models and identify key challenges, including heterogeneous audio utilization, long-horizon temporal degradation, and weak non-speech sound perception, providing insights for future research.
\end{itemize}

\section{Related Work}
\label{sec:related}

\subsection{Proactive Streaming Models}
\label{sec:related_models}

Proactive streaming video understanding requires models to autonomously decide \textit{when} to respond while processing continuous video streams. Existing approaches to this ``when-to-speak'' problem fall into three categories: (1)~\textbf{Token-driven}: the response timing decision is embedded in the autoregressive generation process via special tokens (e.g., EOS, Silence, or Response token), unifying \textit{when} and \textit{what} to speak~\cite{videollm-online,streamo,lion-fs,thinkstream,mmduet2,videollm-eyewo,videollm-mod,proassist,minicpm-o}. (2)~\textbf{Classification-head}: a lightweight, decoupled module explicitly classifies whether to respond at each timestep, separating the timing decision from content generation~\cite{streambridge,dispider,streammind,openhouse,streamagent,em-garde,stride,streamready}. (3)~\textbf{Signal-driven}: response timing is governed by auxiliary signals (e.g., perplexity shifts, or visual scene changes), triggering a response when predefined criteria are met~\cite{livestar,timechat-online}. With triggering mechanisms evolving from simple EOS prediction to reinforcement-learning optimization and sequence denoising, the rapid growth of proactive streaming models makes a comprehensive benchmark that can reliably distinguish a \textit{good} omni-proactive model all the more pressing.

\subsection{Proactive Streaming Video Benchmarks}
\label{sec:related_benchmarks}

We examine existing proactive benchmarks along the three dimensions shown in the blue-shaded columns of \cref{tab:benchmark_comparison}: (1)~\textbf{Omni-modal perception}: whether the benchmark requires audio (speech and non-speech sound) to complete tasks, thereby distinguishing omni-modal models from vision-only ones. (2)~\textbf{Proactive responding}: whether the model autonomously decides when to respond, rather than being polled or queried at preset time points. (3)~\textbf{Diverse video understanding tasks}: how many of the 6 basic video understanding capabilities are covered.

\textbf{StreamingBench-Pro}~\cite{streamingbench} contains 250 purely visual questions from sports/gaming videos. The evaluator polls the model every second and terminates upon the first positive response, meaning each question triggers at most one response. All questions are visual-condition-based, requiring no audio. It covers only Alert (1/6 capabilities).
\textbf{OVO-Bench-Pro}~\cite{ovobench}, despite being labeled ``proactive'', is effectively multi-point static QA. OVO-Bench-Pro queries the model at several preset time points, remaining essentially offline. Since the model never initiates responses on its own, proactive responding is not evaluated. It covers Counting and weak Monitoring (2/6), again without audio involvement.
\textbf{OmniMMI-Pro}~\cite{omnimmi} is the only existing benchmark that supports genuine proactive responding: its Proactive Alert subset lets the model freely decide when to respond in an online streaming setting, and $\sim$35\% of questions require understanding speech content. However, this subset allows only a single response per question, leaving multi-trigger decision-making untested. Moreover, speech is the only audio modality involved, and non-speech sound is entirely absent. Its Proactive Turn-Taking subset is a classification task unrelated to video understanding. Overall, only Alert (1/6) is covered.

In summary, no existing benchmark simultaneously satisfies all three criteria (see \cref{tab:benchmark_comparison}): none involves non-speech sound, only OmniMMI-Pro supports proactive responding (limited to single-trigger), and at most 2/6 capabilities are covered. \benchname systematically addresses these gaps: 84\% of samples require or benefit from audio (both speech and non-speech sound), online evaluation supports multiple responses per question with penalties for over-triggering, and 9 sub-tasks comprehensively cover all 6 capabilities.


\section{Proposed Benchmark}
\label{sec:benchmark}

This section describes \benchname in two parts. \cref{sec:construction} presents how the benchmark is constructed, including the task taxonomy, data sources, automated generation pipeline, human quality control, and resulting dataset statistics. \cref{ssec:eval-protocol} describes how to use the benchmark, detailing the dual-mode evaluation protocol and associated metrics.

\subsection{Construction of \benchname}
\label{sec:construction}

\subsubsection{Task Taxonomy}
\label{sec:taxonomy}

We categorize tasks by cognitive ability into three levels, namely \textit{Perception}, \textit{Comprehension}, and \textit{Reasoning}, with increasing difficulty. This yields 9 sub-tasks and 2,700 evaluation samples in total, see \cref{fig:benchmark} for the complete taxonomy.

\textbf{Instant Event Alert (Event-Alert)} [Perception].
The user specifies a concrete instantaneous event (e.g., a doorbell ringing or a referee's whistle), and the model must issue an alert the moment it occurs. The core challenge is low-latency signal-level pattern matching.

\textbf{Real-time State Monitoring (State-Monitor)} [Perception].
The model continuously monitors a discrete state variable and proactively reports whenever a transition occurs, stating \textit{from} and \textit{to} which state (e.g., ``monitor the dashboard temperature and report changes''). By contrast to Event-Alert, State-Monitor requires sustained perception combined with short-term memory.

\textbf{Snapshot Counting (Snap.-Count)} [Perception].
The model must autonomously detect trigger events (audio or visual) in the video stream and, upon each trigger, count the designated targets currently present in the scene (e.g., ``every time the referee blows the whistle, count the players on the field''). The core challenge lies in coupling event detection with instantaneous counting.

\textbf{Explicit Target Grounding (Target-Ground)} [Perception].
The user specifies a target category, and the model proactively provides its spatial coordinates when the target appears (e.g., ``when a white cat appears, give its coordinates''), combining proactive detection with spatial localization.

\textbf{Event Narration (Event-Narr.)} [Comprehension].
The model performs real-time narration of the streaming content (e.g., ``provide live commentary for this football match''), autonomously determining when noteworthy events occur and proactively producing descriptions. This task demands continuous semantic understanding together with decisions on output timing and granularity.

\textbf{Cumulative Counting (Cum.-Count)} [Comprehension].
The model incrementally counts occurrences of a specified event across time (e.g., ``count how many times the host says `thank you'\,''), demanding persistent tracking and count updates over extended horizons, unlike the snapshot counting in Snap.-Count.

\textbf{Semantic Condition Alert (Cond.-Alert)} [Comprehension].
The user provides an abstract condition (e.g., ``alert me when someone uses inappropriate language''), and the model must understand its semantics and issue an alert when satisfied. Unlike Event-Alert, the trigger is an abstract concept requiring semantic reasoning rather than a concrete physical signal.

\textbf{Deduplicated Counting (Dedup.-Count)} [Reasoning].
The model counts the number of \textit{distinct} targets throughout the video (e.g., ``how many different persons appeared in total?''). Unlike Cum.-Count, Dedup.-Count requires determining whether a currently observed target has appeared before, involving cross-temporal re-identification.

\textbf{Sequential Step Instruction (Step-Inst.)} [Reasoning].
The model assesses the user's current progress in a procedural task and proactively provides next-step guidance at the right moment (e.g., ``teach me to cook scrambled eggs with tomatoes and tell me the next step''). This jointly demands temporal understanding, visual state estimation, and knowledge-based reasoning.

Collectively, these 9 sub-tasks cover 6 basic video understanding capabilities (Alert, Monitoring, Grounding, Counting, Narration, and Prediction), as illustrated in \cref{fig:benchmark}.

\subsubsection{Source Video Collection}
\label{sec:source}

Source videos were drawn from the test sets of two public datasets: LongVALE~\cite{longvale} and COIN~\cite{coin}. LongVALE is a high-quality audio-visual correlation dataset containing diverse long-form videos spanning daily life, sports, and news broadcasts, from which we collected 1,171 videos to supply material for most sub-tasks. However, LongVALE contains limited instructional videos with clear procedural steps as required by the Step-Inst.\ sub-task. To address this, we randomly sampled 600 videos from the COIN test set, which provides comprehensive coverage of step-by-step instructional content. In total, we obtained 1,771 source videos for subsequent QA generation.

\subsubsection{Automated QA Generation}
\label{sec:qa_gen}

\textbf{Dense Captioning}.
For each source video, we employed Gemini 3 Flash to generate temporally aligned multi-modal dense captions with start and end timestamps for each segment. Each segment was described along four fields: \textit{caption} (event omni-summary), \textit{visual} (scene details), \textit{audio} (ambient sounds and music), and \textit{speech} (transcribed spoken content).

\textbf{QA Pair Synthesis}.
We fed both the original video and the dense captions to Gemini 3 Flash, along with a task-specific prompt, to synthesize structured QA samples. Each sample contains the following fields: (1)~\textit{question}: a natural-language standing instruction issued at the start of the video; (2)~\textit{trigger time}: the precise timestamp at which the model should respond; (3)~\textit{response}: the expected proactive output at each trigger time; (4)~\textit{trigger modality}: the modality required to detect the trigger (visual / sound / speech, or combinations); and (5)~\textit{audio dependency}: whether audio is required, helpful, or unnecessary to answer the question.

The generation process adhered to three principles. For \textit{question} design, we adopted an \textbf{audio-first} strategy: prioritize events from the \textit{audio} and \textit{speech} fields, resorting to visual events only as a supplement. For \textit{response} generation, we enforced a \textbf{streaming constraint}: responses must only reference information available up to the trigger time, without using any future video content. For \textit{trigger time} accuracy, we treated the \textbf{video as ground truth}: the dense caption served as a reference, but all timestamps were verified against the actual video content.

Following this pipeline, we automatically generated approximately 1,000 samples per sub-task, yielding 9,000 raw QA samples in total. The full prompt templates for dense captioning and QA generation are provided in the appendix.

\subsubsection{Human Quality Control}
\label{sec:qc}

The auto-generated data underwent two rounds of human review. In the first round, 9 annotators each reviewed one sub-task using a dedicated tool, verifying question naturalness, trigger time accuracy (the precise moment when the trigger event has fully occurred), response faithfulness (free of hallucination), and modality annotation correctness. Annotators revised flawed samples or discarded those of unacceptable quality. In the second round, annotators swapped sub-tasks for cross-validation, ensuring consistent standards across tasks. After both rounds, approximately 30\% of samples were retained, yielding 2,700 samples across 1,262 videos.

\subsubsection{Dataset Statistics}
\label{sec:stats}

\begin{figure}[t]
  \centering

  \begin{minipage}[b]{0.42\linewidth}
    \centering
    \includegraphics[width=\linewidth]{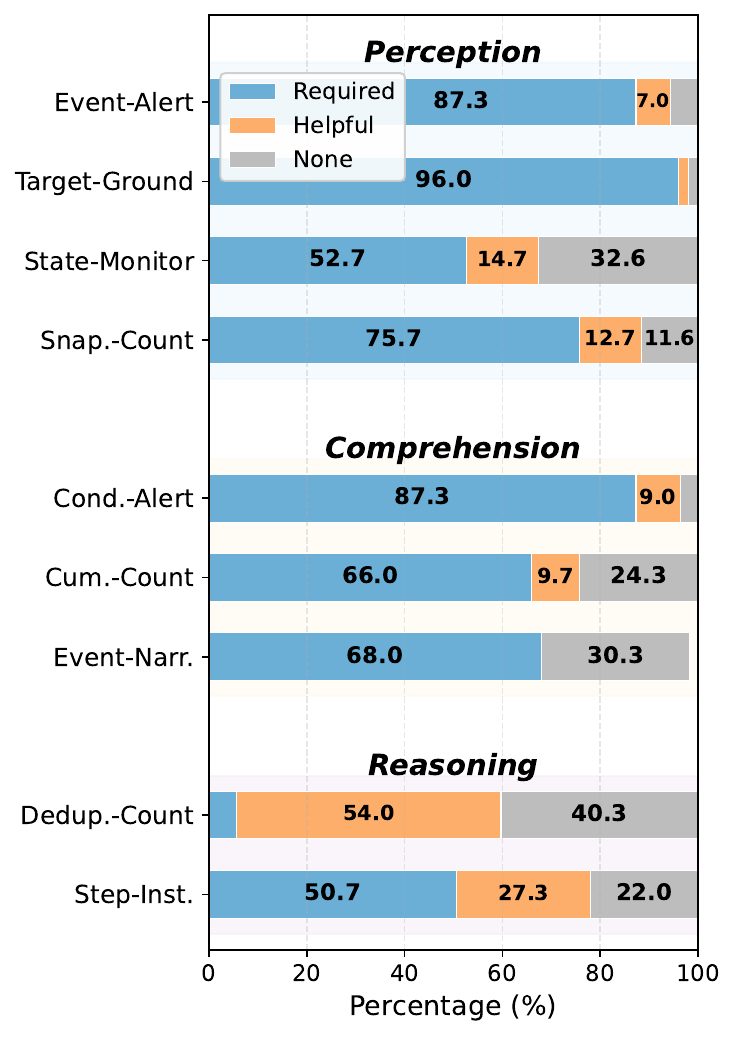}
    \subcaption{Audio dependency per sub-task}
    \label{fig:audio_dep}
  \end{minipage}%
  \hfill
  \begin{minipage}[b]{0.58\linewidth}

    \begin{minipage}[b]{0.48\linewidth}
      \centering
      \includegraphics[width=\linewidth]{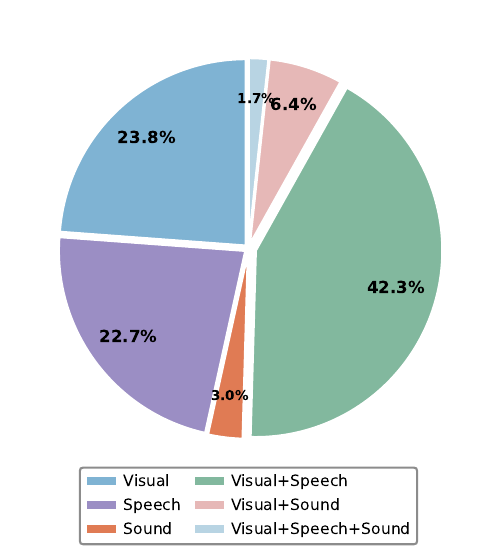}
      \subcaption{Trigger modality ratio}
      \label{fig:trigger_modality}
    \end{minipage}%
    \hfill
    \begin{minipage}[b]{0.48\linewidth}
      \centering
      \includegraphics[width=\linewidth]{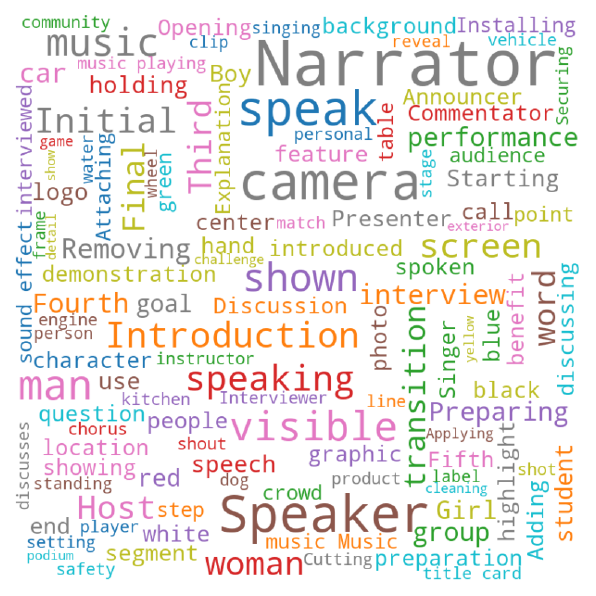}
      \subcaption{Trigger event word cloud}
      \label{fig:wordcloud}
    \end{minipage}

    \vspace{0.3cm}

    \begin{minipage}[b]{\linewidth}
      \centering
      \includegraphics[width=\linewidth]{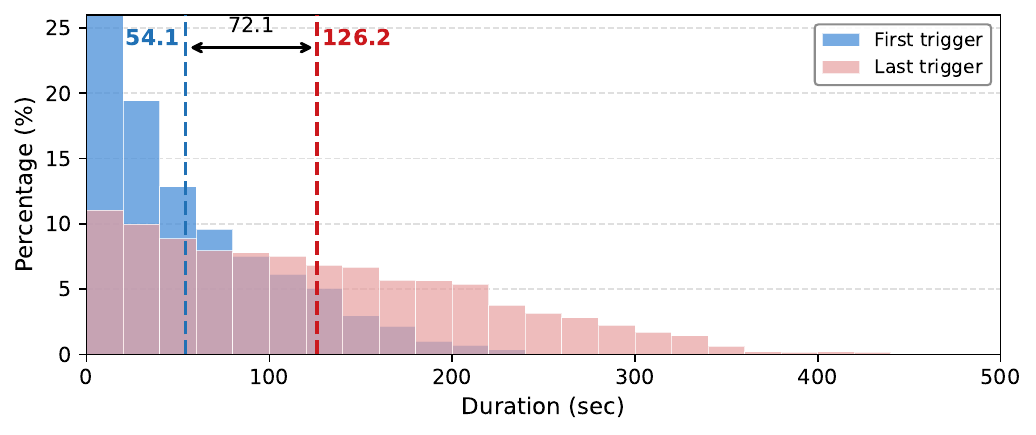}
      \subcaption{First vs.\ last trigger time distribution}
      \label{fig:trigger_time}
    \end{minipage}

  \end{minipage}

  \caption{Dataset statistics of \benchname.}
  \label{fig:stats}
\end{figure}

\cref{fig:stats} visualizes the key distributional properties of \benchname from four perspectives. \cref{fig:audio_dep} shows the audio dependency per sub-task: tasks such as Target-Ground and Event-Alert are almost entirely audio-triggered, whereas Dedup.-Count relies primarily on vision. \cref{fig:trigger_modality} breaks down the trigger modality composition, revealing that visual+speech is the dominant type and nearly half of all triggers exhibit cross-modal characteristics, which ensures the benchmark can differentiate omni models from vision-only counterparts. \cref{fig:wordcloud} displays the diversity of trigger events via a word cloud, showing broad coverage of both audio-related and visual-related triggers. \cref{fig:trigger_time} depicts the distribution of first and last trigger times: the average first trigger occurs at 54.1\,s and the last at 126.2\,s, with a 72.1\,s gap between them, indicating that models must sustain attention across extended durations to achieve high performance.

\subsection{Use of \benchname}
\label{ssec:eval-protocol}

\subsubsection{Evaluation Protocol}
\label{sec:eval_settings}

We design two complementary evaluation modes.

\textbf{Probe mode} is compatible with any VLM and does not require streaming capability. For each ground-truth trigger, the evaluator queries the model twice: a \textit{pre-probe} ($-5$ to $-2$\,s before the trigger) and a \textit{post-probe} ($0$ to $+3$\,s after). In both cases, the model receives the cumulative video frames $[0, t]$ up to the query time and returns a single response. A pre-probe expects a negative answer (the event has not yet occurred), while a post-probe expects the correct task-specific answer. All sub-tasks use dedicated prompt templates that constrain outputs into structured formats (e.g., YES/NO, a single integer, a state name, or a letter choice), including Event-Narr.\ and Step-Inst.\ which are converted into multiple-choice questions. Correctness is determined by exact match for all tasks.

For Probe mode, we report \textbf{Accuracy}. A ground-truth trigger is counted as correct only when \textit{both} its pre-probe and post-probe are answered correctly. The final score is the proportion of correctly answered triggers over all triggers in the benchmark.

\textbf{Online mode} targets streaming models. The model receives the user instruction at the start of the video, then processes subsequent frames one by one together with its own dialogue history, and autonomously decides when to produce a response. No additional queries are issued during the stream. For most sub-tasks, correctness is verified via exact match on structured outputs (e.g., integer count, YES/NO). For open-ended generation tasks (i.e., Event-Narr.\ and Step-Inst.) where output cannot be constrained into a fixed format, we employ Gemini-3-Flash as an LLM judge to score each prediction against the ground truth on a 1--5 scale; a score $\geq$3 is considered correct.

For Online mode, we report \textbf{F1}. Model responses are matched to ground-truth triggers via greedy temporal alignment with a tolerance of $\pm$3\,s. A match is considered valid only if the response is also content-correct. Precision is the fraction of model responses that are validly matched, recall is the fraction of ground-truth triggers that are validly matched, and F1 is their harmonic mean.

\textbf{Model applicability}. Probe mode is applicable to any vision-language model, regardless of whether it supports streaming inference. Online mode requires models with native streaming capability, i.e., models that can process video frame-by-frame and autonomously emit responses. Models that support both paradigms (e.g., MiniCPM-o 4.5) can be evaluated under both modes, while non-streaming models (e.g., InternVL3.5, Qwen3-VL) are evaluated in Probe mode only.

\section{Experiments}
\label{sec:experiments}

\subsection{Experimental Settings}
\label{sec:settings}

\textbf{Evaluated Models}.
We evaluate 11 representative models spanning two evaluation modes. In Probe mode, we assess 9 models: five open-source omni-modal models (Qwen2.5-Omni~\cite{qwen2.5-Omni} 7B, Qwen3-Omni~\cite{qwen3-omni} 30B, video-SALMONN~2+~\cite{video-salmonn2} 7B, VideoLLaMA2.1-AV~\cite{videollama2} 7B, and Phi-4-multimodal~\cite{phi4-multimodal} 14B), two open-source vision-only models (InternVL3.5~\cite{internvl3.5} 8B and Qwen3-VL~\cite{qwen3-vl} 8B), one proprietary omni-modal model (Gemini-3-Flash), and MiniCPM-o~4.5~\cite{minicpm-o} (9B) as the best-performing online model for cross-mode comparison. In Online mode, we evaluate 3 streaming models: MiniCPM-o~4.5 (omni-modal), MMDuet2~\cite{mmduet2} (3B, vision-only), and LiveStar~\cite{livestar} (8B, vision-only). This selection covers multiple contrast dimensions: omni-modal vs.\ vision-only, open-source vs.\ proprietary, and 3B to 30B parameter scales.

\textbf{Implementation Details}.
All models uniformly sample input video at 1\,fps. All open-source model inference is conducted on NVIDIA A800 80GB GPUs. Greedy decoding is used for all open-source models with a maximum generation length of 512 tokens.

\subsection{Using \benchname for Assessing Overall Model Capability}
\label{sec:main_results}

\begin{table*}[htbp!]
\caption{\textbf{Main results}. Per mode, the best and second-best results are shown in \textbf{bold} and \underline{underline}. 
}
\centering
\setlength{\tabcolsep}{4pt}
\renewcommand{\arraystretch}{1.4}
\resizebox{\linewidth}{!}{
\begin{tabular}{@{}lr rrrr rrr rr r@{}}
\toprule
& & \multicolumn{4}{c}{\textbf{Perception}} & \multicolumn{3}{c}{\textbf{Comprehension}} & \multicolumn{2}{c}{\textbf{Reasoning}} & \\
\cmidrule(lr){3-6} \cmidrule(lr){7-9} \cmidrule(lr){10-11}
\multirow{-2}{*}{\textbf{Model}} & \multirow{-2}{*}{\textbf{Params}} & \makecell{Event-\\Alert} & \makecell{Target-\\Ground} & \makecell{State-\\Monitor} & \makecell{Snap.-\\Count} & \makecell{Cond.-\\Alert} & \makecell{Cum.-\\Count} & \makecell{Event-\\Narr.} & \makecell{Dedup.-\\Count} & \makecell{Step-\\Inst.} & \multirow{-2}{*}{\textbf{Mean}} \\
\midrule
\rowcolor{gray!10}
\multicolumn{12}{@{}l}{\textit{Probe-mode evauation (metric: Accuracy):}} \\
InternVL3.5~\cite{internvl3.5} & 8B & 4.8 & 2.4 & 7.2 & 6.0 & 9.3 & 5.3 & 33.0 & 21.3 & 20.0 & 12.1 \\
VideoLLaMA2.1-AV~\cite{videollama2} & 7B & 21.8 & 1.5 & 5.6 & 2.3 & \textbf{24.1} & 4.1 & 27.8 & 9.3 & 14.0 & 12.3 \\
Phi-4-multimodal~\cite{phi4-multimodal} & 14B & 13.7 & 5.1 & 11.5 & 6.0 & 13.8 & 2.0 & 31.0 & 16.1 & 16.9 & 12.9 \\
Qwen3-VL~\cite{qwen3-vl} & 8B & 7.5 & 2.8 & 18.2 & 13.1 & 9.0 & 11.2 & \underline{55.8} & 31.8 & 25.8 & 19.5 \\
Qwen2.5-Omni~\cite{qwen2.5-Omni} & 7B & 35.4 & 8.5 & 8.6 & 18.0 & \underline{18.5} & 9.0 & 49.1 & 15.3 & 18.2 & 20.1 \\
video-SALMONN 2+~\cite{video-salmonn2} & 7B & \underline{37.2} & \textbf{18.1} & 12.3 & \underline{24.7} & 17.6 & 11.5 & 41.3 & 20.3 & 15.6 & 22.1 \\
Qwen3-Omni~\cite{qwen3-omni} & 30B & 21.5 & 10.4 & 18.3 & 19.3 & 9.9 & 15.3 & 46.8 & 30.0 & \underline{31.6} & 22.6 \\
MiniCPM-o 4.5~\cite{minicpm-o} & 9B & 18.2 & \underline{16.4} & \underline{28.2} & \textbf{28.0} & 9.8 & \underline{27.9} & 45.9 & \underline{32.5} & 25.8 & \underline{25.8} \\
Gemini-3-Flash & -- & \textbf{38.2} & 12.1 & \textbf{35.0} & 21.0 & 12.8 & \textbf{42.7} & \textbf{86.4} & \textbf{39.6} & \textbf{76.3} & \textbf{40.4} \\
\midrule
\rowcolor{gray!10}
\multicolumn{12}{@{}l}{\textit{Online-mode evaluation (metric: F1):}} \\
LiveStar~\cite{livestar} & 8B & 9.7 & 0.8 & 0.0 & 0.0 & 14.7 & 0.0 & 1.6 & 0.0 & 6.0 & 3.6 \\
MMDuet2~\cite{mmduet2} & 3B & \underline{12.5} & \underline{5.3} & \underline{14.9} & \underline{11.2} & \underline{21.4} & \underline{5.3} & \underline{3.7} & \underline{12.7} & \textbf{14.7} & \underline{11.3} \\
MiniCPM-o 4.5~\cite{minicpm-o} & 9B & \textbf{44.2} & \textbf{13.9} & \textbf{24.3} & \textbf{21.2} & \textbf{33.1} & \textbf{16.4} & \textbf{6.9} & \textbf{20.5} & \underline{7.9} & \textbf{20.9} \\
\bottomrule
\end{tabular}
}
\label{tab:main_results}
\end{table*}

\cref{tab:main_results} presents the main results. Overall, current models achieve modest performance, confirming that omni-proactive streaming video understanding remains a challenging open problem. We highlight four observations. \textbf{(1)} Gemini-3-Flash attains 40.4\% average accuracy, nearly double the best open-source model (22.1\%), indicating a substantial capability gap between proprietary and open-source systems. \textbf{(2)} On audio-dependent tasks (e.g., Event-Alert), omni-modal models surpass vision-only counterparts by over 30 points, confirming that audio perception is critical and vision alone is insufficient for these tasks. \textbf{(3)} Online mode is considerably harder: MiniCPM-o~4.5 reaches only 20.9\% F1, with severe degradation on generation-intensive tasks (Event-Narr.\ 6.9\%, Step-Inst.\ 7.9\%), exposing the coupled challenge of deciding \textit{when to speak} and producing correct content simultaneously. \textbf{(4)} Reasoning-level tasks exhibit the largest capability gap (Step-Inst.: 76.3 for Gemini vs.\ 31.6 for the best open-source), suggesting that multi-step causal inference remains the most difficult capability to acquire.

\subsection{Using \benchname for Disentangling Modality Contributions}
\label{sec:analysis_a}

\begin{table}[t]
\caption{\textbf{Impact of input information for OmniLLMs}. We conduct experiments across three input configurations: audio-only, video-only, and video with original audio. The $\Delta \uparrow$ in the Mean column of A+V denotes the absolute gain over V.}
\centering
\setlength{\tabcolsep}{3.5pt}
\renewcommand{\arraystretch}{1.3}
\resizebox{\linewidth}{!}{
\begin{tabular}{@{}ll rrrr rrr rr l@{}}
\toprule
& & \multicolumn{4}{c}{\textbf{Perception}} & \multicolumn{3}{c}{\textbf{Comprehension}} & \multicolumn{2}{c}{\textbf{Reasoning}} & \\
\cmidrule(lr){3-6} \cmidrule(lr){7-9} \cmidrule(lr){10-11}
\multirow{-2}{*}{\textbf{Model}} & \multirow{-2}{*}{\textbf{Input}} & \makecell{Event-\\Alert} & \makecell{Target-\\Ground} & \makecell{State-\\Monitor} & \makecell{Snap.-\\Count} & \makecell{Cond.-\\Alert} & \makecell{Cum.-\\Count} & \makecell{Event-\\Narr.} & \makecell{Dedup.-\\Count} & \makecell{Step-\\Inst.} & \multirow{-2}{*}{\textbf{Mean}} \\
\midrule
\multirow{3}{*}{Qwen2.5-Omni~\cite{qwen2.5-Omni}} & A & 33.3 & 5.5 & 7.3 & 2.0 & 16.6 & 2.7 & 35.9 & 0.0 & 15.1 & 13.2 \\
 & V & 9.1 & 4.1 & 6.4 & 10.0 & 8.4 & 5.4 & 40.9 & 16.7 & 19.9 & 13.4 \\
 & A+V & 35.4 & 8.5 & 8.6 & 18.0 & 18.5 & 9.0 & 49.1 & 15.3 & 18.2 & 20.1 (\gain{6.7}$\uparrow$) \\
\midrule
\multirow{3}{*}{video-SALMONN 2+~\cite{video-salmonn2}} & A & 42.4 & 16.4 & 3.6 & 10.0 & 14.7 & 14.2 & 40.0 & 1.5 & 14.4 & 17.5 \\
 & V & 3.0 & 3.6 & 5.0 & 8.0 & 8.0 & 6.9 & 32.7 & 16.8 & 14.8 & 11.0 \\
 & A+V & 37.2 & 18.1 & 12.3 & 24.7 & 17.6 & 11.5 & 41.3 & 20.3 & 15.6 & 22.1 (\gain{11.1}$\uparrow$) \\
\midrule
\multirow{3}{*}{Qwen3-Omni~\cite{qwen3-omni}} & A & 19.7 & 1.8 & 5.0 & 0.0 & 7.4 & 8.2 & 25.0 & 4.1 & 16.8 & 9.8 \\
 & V & 13.3 & 8.4 & 15.4 & 16.8 & 7.6 & 8.5 & 48.9 & 30.0 & 33.3 & 20.2 \\
 & A+V & 21.5 & 10.4 & 18.3 & 19.3 & 9.9 & 15.3 & 46.8 & 30.0 & 31.6 & 22.6 (\gain{2.4}$\uparrow$) \\
\midrule
\multirow{3}{*}{Gemini-3-Flash} & A & 27.3 & 1.8 & 15.0 & 2.0 & 8.0 & 23.7 & 56.8 & 8.1 & 58.7 & 22.4 \\
 & V & 18.2 & 9.1 & 32.3 & 24.0 & 7.5 & 24.7 & 76.8 & 37.1 & 80.2 & 34.4 \\
 & A+V & 38.2 & 12.1 & 35.0 & 21.0 & 12.8 & 42.7 & 86.4 & 39.6 & 76.3 & 40.4 (\gain{6.0}$\uparrow$) \\
\midrule
\multirow{3}{*}{MiniCPM-o 4.5~\cite{minicpm-o}} & A & 42.6 & 11.5 & 6.6 & 7.1 & 18.1 & 3.9 & 3.8 & 1.7 & 2.7 & 10.9 \\
 & V & 14.9 & 8.7 & 23.3 & 16.0 & 15.7 & 7.6 & 3.5 & 27.3 & 7.5 & 13.8 \\
 & A+V & 44.2 & 13.9 & 24.3 & 21.2 & 33.1 & 16.4 & 6.9 & 20.5 & 7.9 & 20.9 (\gain{7.1}$\uparrow$) \\
\bottomrule
\end{tabular}
}
\label{tab:audio_ablation}
\end{table}

\cref{tab:audio_ablation} reports five omni-modal models under audio-only (A), video-only (V), and full audio-visual (A+V) inputs to disentangle modality contributions. Three findings emerge. \textbf{(1)} A+V consistently outperforms either single modality, with gains over V ranging from +2.4 (Qwen3-Omni) to +11.1 (video-SALMONN~2+), confirming that the two modalities provide complementary cues. \textbf{(2)} The relative strength of A vs.\ V is highly task-dependent: on Event-Alert, A dominates V across all models (e.g., 42.4 vs.\ 3.0 for video-SALMONN~2+), whereas on Dedup.-Count and Step-Inst., V substantially outperforms A (e.g., 30.0 vs.\ 4.1 for Qwen3-Omni). \textbf{(3)} Models exhibit divergent modality utilization patterns: video-SALMONN~2+ relies more heavily on audio (A: 17.5 vs.\ V: 11.0), while Qwen3-Omni is predominantly vision-driven (V: 20.2 vs.\ A: 9.8), revealing fundamental differences in audio encoding and multi-modal fusion capabilities.

\subsection{Using \benchname for Evaluating Long-Horizon Perception}
\label{sec:analysis_b}

\begin{figure}[htbp!]
  \centering
  \includegraphics[width=\linewidth]{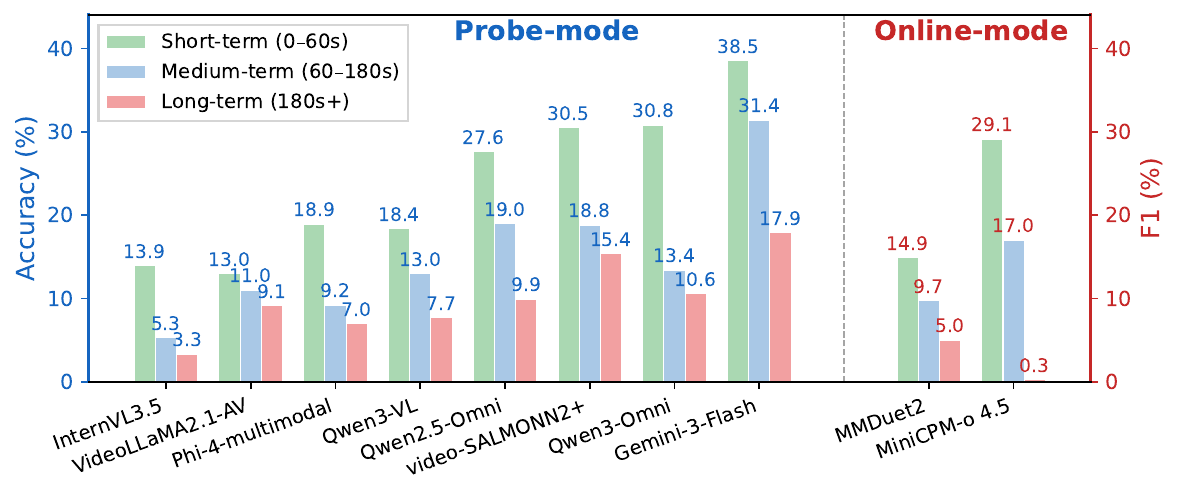}
  \caption{Performance grouped by where the GT trigger is located along the video timeline.}
  \label{fig:temporal_ablation}
\end{figure}

\cref{fig:temporal_ablation} groups performance by where the GT trigger is located along the video timeline: Short-term (0--60\,s), Medium-term (60--180\,s), and Long-term (180\,s+). All models show substantial degradation for later-occurring triggers, retaining on average only 37\% of their Short-term performance at the Long-term. MiniCPM-o~4.5 (Online mode) nearly fails entirely on the Long-term (29.1 $\to$ 0.3), indicating that current streaming models cannot sustain perception over extended video streams. Even Gemini-3-Flash, the strongest offline model, retains only 46\% of its Short-term performance at the Long-term (38.5 $\to$ 17.9), confirming that all models struggle to perceive and respond to events occurring late in long videos.

\subsection{Using \benchname for Identifying Modality Bottlenecks}
\label{sec:analysis_c}

\begin{wrapfigure}{r}{0.35\linewidth}
  \centering
  \vspace{-50pt}
  \includegraphics[width=\linewidth]{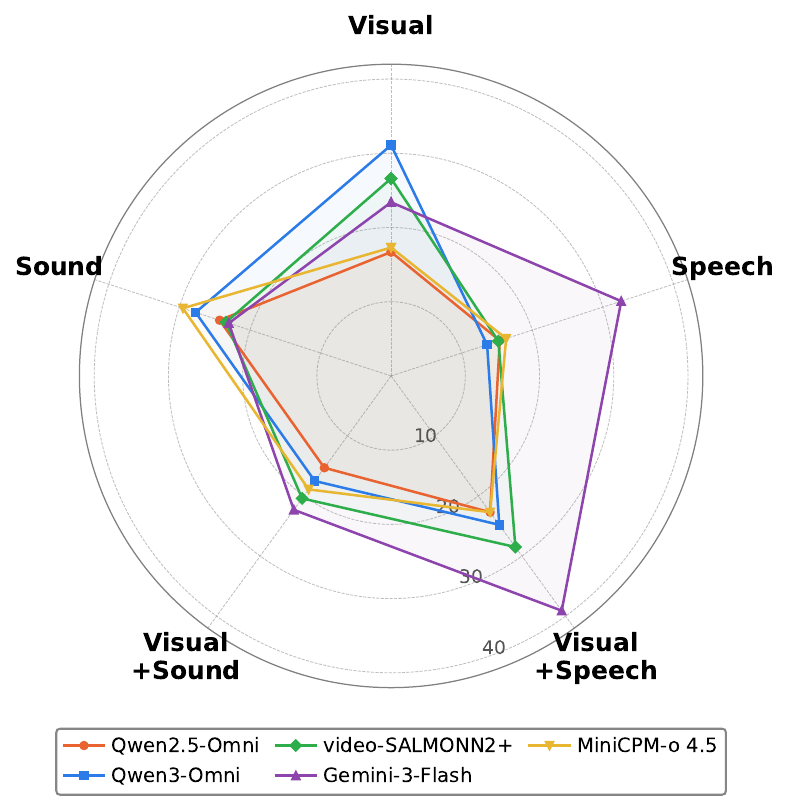}
  \caption{Performance breakdown by the modality signals required to perceive the trigger event.}
  \label{fig:radar}
  \vspace{-50pt}
\end{wrapfigure}

\cref{fig:radar} breaks down performance by the modality signals required to perceive each trigger event: visual only, speech, visual+speech, and visual+sound (non-speech audio). Gemini-3-Flash dominates on speech and visual+speech triggers (32.6 and 39.1, respectively), yet falls behind Qwen3-Omni on pure visual triggers (23.4 vs.\ 31.1), indicating that its advantage stems primarily from speech comprehension rather than visual perception. All models perform weakest on visual+sound triggers (15.3--22.3), revealing that perceiving and utilizing non-speech audio (e.g., environmental sounds, sound effects) remains a shared bottleneck.

\section{Conclusions}
\label{sec:conclusion}

We have presented \benchname, the first comprehensive benchmark for omni-proactive streaming video understanding, comprising 2,700 human-verified samples across 9 sub-tasks and 3 cognitive levels with 84\% audio dependency, together with a dual-mode evaluation protocol (Probe and Online) that enables joint assessment of omni-modal perception, proactive responding, and diverse video understanding tasks. Evaluation of 11 representative models reveals that: (1)~a substantial gap persists between proprietary and open-source systems (40.4\% vs.\ 22.6\%), particularly on reasoning-level tasks; (2)~audio and video provide complementary cues, yet models exhibit divergent modality utilization patterns; (3)~all models struggle to perceive events occurring late in long videos, with online streaming models nearly failing beyond 180\,s; and (4)~non-speech audio perception remains the weakest dimension across all models. We hope \benchname serves as a useful testbed for driving progress toward genuine omni-proactive streaming video understanding.

\begin{ack}
This research was supported by NSFC (No. 62576348), BJNSF (No. L254039) and Tencent WeChat Rhino-Bird Focused Research Program.
\end{ack}

\bibliographystyle{plainnat}
\bibliography{references}

\appendix
\section{More Experimental Results}
\label{app:results}

\subsection{Tolerance Window Ablation}

\begin{figure}[h]
  \centering
  \includegraphics[width=0.75\linewidth]{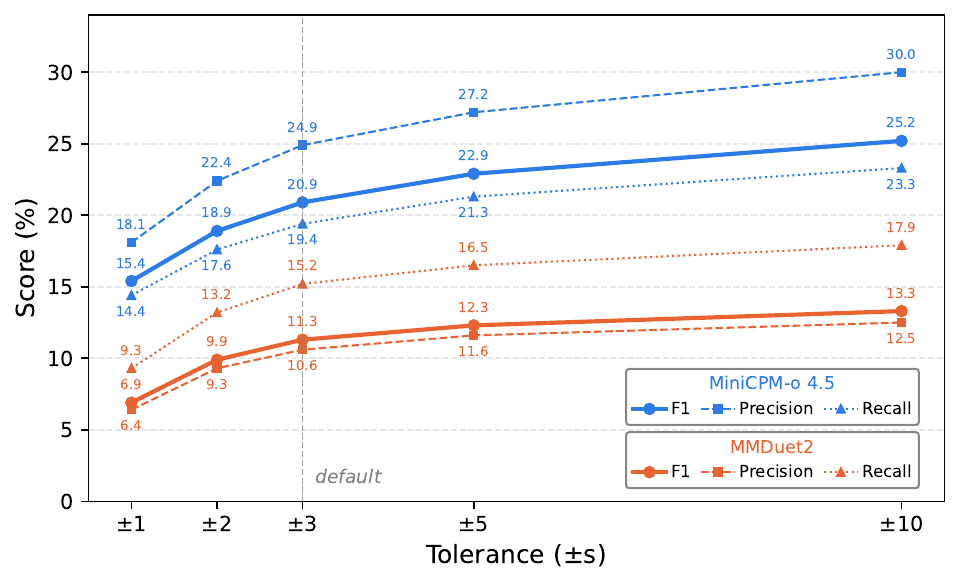}
  \caption{\textbf{Tolerance window ablation (Online mode).} Performance of online-mode models under varying temporal matching tolerances.}
  \label{fig:tolerance_ablation}
\end{figure}

\cref{fig:tolerance_ablation} shows the effect of varying the temporal matching tolerance on joint\_F1 for three online-mode models (MiniCPM-o~4.5, MMDuet2, and LiveStar). The tolerance window ranges from $\pm$1\,s to $\pm$5\,s. We adopt $\pm$3\,s as the default in all Online-mode evaluations.




\section{More Details of Data Construction}
\label{app:data}

\subsection{Dense Captioning Prompt}
\label{app:dense_caption_prompt}

We first generate temporally aligned multi-modal dense captions for each source video using Gemini 3 Flash. The following prompt template is used, where \texttt{\{duration\_mmss\}}, \texttt{\{duration\_sec\}}, and \texttt{\{suggested\_segments\}} are filled per video.

\begin{tcolorbox}[colback=gray!5, colframe=gray!80, title={\textbf{Dense Captioning Prompt}}, fonttitle=\small\color{white}, colbacktitle={rgb,255:red,64;green,64;blue,64}, boxrule=0.5pt, arc=2pt, breakable]
\small
\begin{verbatim}
You are an expert video analyst producing annotations for a video
understanding dataset. Watch this video carefully — pay close
attention to the visual content, sounds/music, and speech/dialogue
simultaneously.

This video is {duration_mmss} long ({duration_sec:.0f} seconds).

Produce a dense temporal caption — divide the video into
fine-grained segments. These captions will be used downstream to
generate question-answer pairs about the video, so they must
contain enough detail to support questions about what happened,
when, why, what changed, and what can be inferred.

Segment guidelines:
- Each segment should be roughly 5-30 seconds long. No segment
  should exceed 45 seconds.
- For a {duration_sec:.0f}-second video, produce approximately
  {suggested_segments} segments.
- Start a new segment when there is: a scene/location change, a
  new activity or action, a topic shift in speech, a new person
  appearing or leaving, an audio event (music change, sound
  effect, silence), or any notable visual change.

For each segment, provide:

1. caption: A detailed, information-dense paragraph integrating
   visual, audio, and speech into one coherent description.
   Include: Who (appearance, actions), What (objects, text),
   Action (specific verbs, direction), Change (differences from
   previous segment), Audio-visual correlation.

2. visual: Exhaustive visual details — scene, lighting, colors,
   objects, people, camera work, on-screen text verbatim.

3. audio: Precise sound description — music (genre, tempo,
   instruments), sound effects, ambient sounds, voice quality.
   Note onset and cessation of sounds.

4. speech: Detailed summary of what is said — key claims, names,
   numbers, facts. If no speech, write "None".

Return a JSON array:
[
  {
    "start": "MM:SS",
    "end": "MM:SS",
    "caption": "...",
    "visual": "...",
    "audio": "...",
    "speech": "..."
  }
]

Rules:
- Segments must cover the entire video from 00:00 to
  {duration_mmss} with no gaps or overlaps.
- Timestamps in MM:SS format.
- Be maximally specific — names, colors, counts, positions.
- Return raw JSON only, no markdown.
\end{verbatim}
\end{tcolorbox}

\subsection{QA Generation Prompts}
\label{app:prompts}

We provide condensed prompt templates used to generate QA pairs for each sub-task. All prompts share the system preamble ``You are an expert at constructing QA benchmark data for evaluating proactive omni-modal assistants'' and are fed to Gemini 2.5 Flash together with the source video and dense captions. Full prompts are available in the code repository.

\begin{tcolorbox}[colback=gray!5, colframe=gray!80, title={\textbf{Instant Event Alert (IEA)}}, fonttitle=\small\color{white}, colbacktitle={rgb,255:red,64;green,64;blue,64}, boxrule=0.5pt, arc=2pt, breakable]
\small
\begin{verbatim}
## Task
"Instant Event Alert" — user gives a standing instruction
(e.g., "Let me know when the kettle whistles"); assistant
watches/listens and proactively responds at the right
moment(s) when the target event occurs.

## Input
1. Original video (ground truth).
2. Timestamped dense caption (supplementary reference).
Duration: {duration_mmss} ({duration_sec:.0f}s).

## Steps

1. Choose an event (AUDIO-FIRST priority):
   - Audio-required (best): ONLY detectable by listening
     (doorbell, whistle, spoken phrase, alarm, glass break)
   - Audio-helpful: visible but audio confirms
   - Visual-only (last resort): no meaningful audio

2. Write the question:
   One natural standing instruction at 00:00. Must sound
   like a real person talking to a smart assistant.
   No spoilers, no timestamps, everyday language.

3. Write response(s) — one per event occurrence:
   - State what happened, briefly and naturally.
   - Include accurate trigger_time (MM:SS).
   - Conversational tone, not robotic.

4. Classify each response:
   - trigger_type: "visual"|"sound"|"speech"|combined
     (e.g., "visual+sound", "visual+speech")
   - audio_dependency: "required"|"helpful"|"none"
   - trigger_type_reason: brief explanation

## Output (single JSON object, no markdown)
Fields: status, question, question_time ("00:00"),
audio_dependency, responses[] with: trigger_time,
response, trigger_type, trigger_type_reason,
event_description.
If no suitable event: {"status":"skip","reason":"..."}

## Rules
- Raw JSON only. Timestamps MM:SS in
  [00:00,{duration_mmss}].
- Video is ground truth; caption is supplementary.
- Do NOT fabricate events not clearly present.
- Try your best to find an audio-required event.
\end{verbatim}
\end{tcolorbox}

\begin{tcolorbox}[colback=gray!5, colframe=gray!80, title={\textbf{Explicit Target Grounding (ETG)}}, fonttitle=\small\color{white}, colbacktitle={rgb,255:red,64;green,64;blue,64}, boxrule=0.5pt, arc=2pt, breakable]
\small
\begin{verbatim}
## Task
"Explicit Target Grounding" — user specifies a target
object and trigger condition. When trigger fires, assistant
locates target in frame using a 3x3 grid. Tests: (1)
detect trigger in real-time, (2) locate target spatially.

## Input
1. Original video (ground truth).
2. Timestamped dense caption (rough reference).
Duration: {duration_mmss} ({duration_sec:.0f}s).

## Steps

1. Find a trigger-target pair (AUDIO-FIRST):
   TRIGGER: instantaneous, real-time confirmable (NO
   "finishes/ends/completes"), unambiguous (maps to
   precise frame), naturally paired with target.
   OK: whistle->ball, "Maria" called->Maria
   TARGET: fits in ONE grid cell (highest priority).
   Never: full person, large vehicle, close-up face.
   Preferred: small held objects, accessories, buttons,
   logos, license plates, knobs.
   Visible at trigger moment. Single specific object.

2. Write the question:
   One natural instruction at 00:00. Specify BOTH trigger
   and target. Ask for position "in the frame"/"on screen".

3. Write response(s) — default exactly ONE (max 4):
   - Describe trigger event. State target location.
   - position: one of 9 grid cells ("top-left"|
     "top-center"|...|"bottom-right")
   - trigger_time (MM:SS). Under 20 words.

4. Classify: trigger_type, audio_dependency.

## Output (single JSON object, no markdown)
Fields: status, question, question_time, audio_dependency,
responses[] with: trigger_time, response, position,
trigger_type, trigger_type_reason, event_description.
If no pair: {"status":"skip","reason":"..."}

## Rules
- Timestamps MM:SS in [00:00,{duration_mmss}].
- Position and trigger_time must be synchronized.
- Target MUST fit in ONE grid cell (highest priority).
- Trigger and target meaningfully connected.
- Prefer audio triggers. Video is ground truth.
- No hallucination. 1-4 responses; prefer 1.
\end{verbatim}
\end{tcolorbox}

\begin{tcolorbox}[colback=gray!5, colframe=gray!80, title={\textbf{Realtime State Monitor (RSM)}}, fonttitle=\small\color{white}, colbacktitle={rgb,255:red,64;green,64;blue,64}, boxrule=0.5pt, arc=2pt, breakable]
\small
\begin{verbatim}
## Task
"Realtime State Monitor" — user asks assistant to monitor
a discrete, observable state of a main subject and report
whenever it transitions (e.g., sitting->standing,
kitchen->living room, speaking->silent).

## Input
1. Original video (ground truth).
2. Timestamped dense caption (reference).
Duration: {duration_mmss} ({duration_sec:.0f}s).

## Steps

1. Choose a SPECIFIC PHYSICAL DIMENSION:
   Audio scan (MANDATORY) first:
   - Speakers taking turns? -> "who is speaking"
   - Music starts/stops? -> "whether music is playing"
   - Alternating sound sources?
   If any audio dimension works, use it.

   Visual scan (only if no audio):
   State must be: specific (ONE property), discrete,
   about main subject, changes 2+ times, objective.
   GOOD: posture, location/room, orientation, motion,
   object state, worn items, speaker identity, number
   of people, sound source, music present/absent.
   BAD: "activity" (open-ended), speed/volume
   (continuous), music mood (subjective), emotion.

2. Write the question:
   One natural monitoring instruction at 00:00.
   UNAMBIGUOUS. No spoilers. No state value lists.

3. Write responses — ONLY at transitions (2-5):
   - Name previous AND new state (from X -> to Y).
   - trigger_time (MM:SS), after 00:00. Under 15 words.
   - Do NOT report initial state. Chronological.

4. Classify: trigger_type, audio_dependency.
   Include audio_scan field.

## Output (single JSON object, no markdown)
Fields: status, audio_scan, question, question_time,
audio_dependency, responses[] with: trigger_time,
response, trigger_type, trigger_type_reason,
event_description.
If no suitable state: {"status":"skip","reason":"..."}

## Rules
- Timestamps MM:SS in [00:00,{duration_mmss}].
- States must be DISCRETE with clear boundaries.
- Listen first. Prefer audio dimensions.
- Each response names from-state and to-state.
- Do NOT fabricate state changes. Aim 2-5 responses.
\end{verbatim}
\end{tcolorbox}

\begin{tcolorbox}[colback=gray!5, colframe=gray!80, title={\textbf{Snapshot Counting (SC)}}, fonttitle=\small\color{white}, colbacktitle={rgb,255:red,64;green,64;blue,64}, boxrule=0.5pt, arc=2pt, breakable]
\small
\begin{verbatim}
## Task
"Snapshot Counting" — user specifies a trigger moment and
a counting target naturally related to that trigger. When
trigger occurs, assistant counts target entities at that
instant. Tests: (1) detect trigger, (2) count accurately.
Key: trigger and target must be naturally connected.

## Input
1. Original video (ground truth).
2. Timestamped dense caption (reference).
Duration: {duration_mmss} ({duration_sec:.0f}s).

## Steps

1. Find trigger + counting target (LISTEN FIRST):
   Good audio trigger->target pairs (naturally connected):
   - Whistle blows -> count players on field
   - Applause starts -> count performers on stage
   - "everyone ready?" -> count people in room
   - Timer buzzes -> count dishes on counter
   Bad (artificially forced):
   - "Hey" -> count people on sofa (no connection)

   Visual scan (if no audio):
   - New dish placed -> count all dishes
   - Wide shot revealed -> count people

2. Write the question:
   One natural counting instruction at 00:00.
   Specifies BOTH trigger and counting target.
   Natural language. No expected count revealed.

3. Write ONE response at trigger moment:
   - Note trigger occurred. State exact count.
   - count field with integer (for evaluation).
   - trigger_time (MM:SS), after 00:00. Under 15 words.

4. Classify: trigger_type, audio_dependency.
   Include audio_scan field.

## Output (single JSON object, no markdown)
Fields: status, audio_scan, question, question_time,
audio_dependency, responses[] (exactly one) with:
trigger_time, response, count, trigger_type,
trigger_type_reason, event_description.
If no pair: {"status":"skip","reason":"..."}

## Rules
- Timestamps MM:SS in [00:00,{duration_mmss}].
- Exactly ONE response. Must contain count (integer).
- Trigger and target NATURALLY CONNECTED.
- Listen first. Vary targets (not always people).
- Count must be accurate to what's visible at trigger.
- Do NOT fabricate triggers or counts.
\end{verbatim}
\end{tcolorbox}

\begin{tcolorbox}[colback=gray!5, colframe=gray!80, title={\textbf{Semantic Condition Alert (SCA)}}, fonttitle=\small\color{white}, colbacktitle={rgb,255:red,64;green,64;blue,64}, boxrule=0.5pt, arc=2pt, breakable]
\small
\begin{verbatim}
## Task
"Semantic Condition Alert" — user describes a condition
requiring semantic understanding (not keyword/object/sound
detection). Assistant monitors and alerts each time the
condition is met. Tests: (1) understand abstract intent,
(2) map to concrete occurrences via reasoning, (3) alert
at right times. Distinction from IEA: IEA=perception;
SCA=comprehension/judgment.

## Input
1. Original video (ground truth).
2. Timestamped dense caption (reference).
Duration: {duration_mmss} ({duration_sec:.0f}s).

## Steps

1. Find a natural condition (AUDIO-FIRST):
   Must satisfy ALL:
   A. Realistic — a real person would want this alert.
   B. Requires semantic understanding (NOT perception):
      Test: "Could a detector+classifier handle this?"
      BAD: "when audience cheers" (sound classification)
      GOOD: "when speaker provides a statistic as evidence"
   C. Unambiguous (9/10 people flag same moments):
      BAD: "when someone gives advice" (blurry)
   D. Focus on EVENTS/ACTIONS, not linguistic analysis.

2. Write the question:
   One natural monitoring instruction at 00:00.
   Describes condition clearly. No spoilers.

3. Write responses — one per occurrence:
   - State what happened AND why it satisfies condition.
   - Under 25 words. trigger_time (MM:SS), after 00:00.
   - Speech: timestamp = when sentence ends.

4. Classify: trigger_type, audio_dependency.

## Output (single JSON object, no markdown)
Fields: status, question, question_time, audio_dependency,
responses[] with: trigger_time, response, trigger_type,
trigger_type_reason, event_description.
If no condition: {"status":"skip","reason":"..."}

## Rules
- Timestamps MM:SS in [00:00,{duration_mmss}].
- At least 1 response. Each under 25 words.
- Condition must be realistic, unambiguous, and require
  semantic understanding (not perception-level).
- No hallucination — only confident observations.
\end{verbatim}
\end{tcolorbox}

\begin{tcolorbox}[colback=gray!5, colframe=gray!80, title={\textbf{Cumulative Counting (CC)}}, fonttitle=\small\color{white}, colbacktitle={rgb,255:red,64;green,64;blue,64}, boxrule=0.5pt, arc=2pt, breakable]
\small
\begin{verbatim}
## Task
"Cumulative Counting" — user specifies a repeatable event.
Assistant detects each occurrence, keeps running tally,
reports updated cumulative count. Tests: (1) detect each
occurrence, (2) maintain count, (3) report at each event.
Key: events must be discrete and separable.

## Input
1. Original video (ground truth).
2. Timestamped dense caption (reference).
Duration: {duration_mmss} ({duration_sec:.0f}s).
Preferred Category: {preferred_category}

## Event Categories
A — Discrete non-speech sounds: impact, signals,
    instrument hits, animal/body sounds.
B — Speech acts: questions, instructions, jokes,
    laughter bursts. Each = one complete act.
C — Word/phrase repetitions: meaningful word said
    multiple times. NOT: function words.
D — Repeating visual actions: exercise reps, chops,
    spins. NOT: continuous stirring/swaying.

## Steps

1. Find event (try {preferred_category} first):
   Discrete/separable (10 people agree), repeats 3+
   times (aim 3-8), non-overlapping, unambiguous.

2. Write the question:
   One natural counting instruction at 00:00.
   Specifies event clearly. No count revealed.

3. Write responses — one per occurrence:
   - Natural notification (NOT "Count: X").
   - count: cumulative integer (1, 2, 3, ...).
   - trigger_time (MM:SS), after 00:00. Under 20 words.
   - Chronological, incrementing by exactly 1.

4. Classify: trigger_type, audio_dependency.
   Include chosen_category field.

## Output (single JSON object, no markdown)
Fields: status, chosen_category, question, question_time,
audio_dependency, responses[] with: trigger_time,
response, count, trigger_type, trigger_type_reason,
event_description.
If <3 occurrences: {"status":"skip","reason":"..."}

## Rules
- Timestamps MM:SS in [00:00,{duration_mmss}].
- At least 3 responses. Each under 20 words.
- count increments by exactly 1 chronologically.
- Events MUST be discrete (10 people agree on count).
- No hallucination — only events clearly present.
\end{verbatim}
\end{tcolorbox}

\begin{tcolorbox}[colback=gray!5, colframe=gray!80, title={\textbf{Event Narration (EN)}}, fonttitle=\small\color{white}, colbacktitle={rgb,255:red,64;green,64;blue,64}, boxrule=0.5pt, arc=2pt, breakable]
\small
\begin{verbatim}
## Task
"Event Narration" — user specifies a narration focus;
assistant provides concise, factual updates at natural
breakpoints. Tests: (1) real-time comprehension, (2)
identifying breakpoints, (3) accurate summaries.
NOT open-ended — always constrained to a specific focus.

## Input
1. Original video (ground truth).
2. Timestamped dense caption (reference).
Duration: {duration_mmss} ({duration_sec:.0f}s).

## Streaming Constraint
Real-time — NO future knowledge. Each update describes
only what happened UP TO that point. NEVER use
"concludes/climax/final/wraps up/ends with."

## Steps

1. Find a natural narration focus (satisfy ALL):
   A. Specific and constrained (NOT "describe everything")
   B. Multiple natural breakpoints (3+ stages).
   C. Grounded in observable, verifiable facts.
   D. Realistic. E. Integrates visual AND audio.

2. Write the question:
   One natural instruction at 00:00. Specifies focus,
   implies ongoing updates. No spoilers.

3. Write responses — one per breakpoint (aim 3-6):
   - Factual summary since last update, within focus.
   - Specific verifiable details (names, quantities).
   - trigger_time (MM:SS). Under 40 words.
   - Chronological. Each adds NEW information.
   - Distributed across video (max 2 in first quarter).

4. Classify: trigger_type, audio_dependency.

## Output (single JSON object, no markdown)
Fields: status, question, question_time, audio_dependency,
responses[] with: trigger_time, response, trigger_type,
trigger_type_reason, event_description.
If no focus: {"status":"skip","reason":"..."}

## Rules
- Timestamps MM:SS in [00:00,{duration_mmss}].
- 3-6 responses. Each under 40 words. trigger_time>00:00.
- Stay within focus. Every fact verifiable from video.
- Narrate at natural breakpoints, not fixed intervals.
- NO "conclusion/finale/climax" language.
- Accuracy over coverage. Video is ground truth.
\end{verbatim}
\end{tcolorbox}

\begin{tcolorbox}[colback=gray!5, colframe=gray!80, title={\textbf{Deduplicated Counting (DC)}}, fonttitle=\small\color{white}, colbacktitle={rgb,255:red,64;green,64;blue,64}, boxrule=0.5pt, arc=2pt, breakable]
\small
\begin{verbatim}
## Task
"Dedup Counting" — user specifies a target category.
Assistant detects each unique new target, maintains tally
of distinct targets, reports — ignoring re-appearances.
Tests: (1) detect targets, (2) identity tracking, (3)
only count genuinely new targets. Distinction from CC:
CC counts events; DC counts unique entities.

## Input
1. Original video (ground truth).
2. Timestamped dense caption (rough reference).
Duration: {duration_mmss} ({duration_sec:.0f}s).

## Steps

1. Check required appear/disappear/reappear pattern:
   - Targets appear at spread-out times?
   - At least one disappears and reappears later?
   - At least 3 unique targets?
   If no reappear pattern, return skip.

2. Find target category (must satisfy ALL):
   - Distinct identities. Appear-disappear-reappear.
   - 3+ targets, min 15s span. Unambiguous (9/10 agree).
   - Precisely scoped with qualifier when noisy:
     GOOD: "people interviewed on camera", "products
     picked up and demonstrated"
     BAD: "different scenes" (vague)

3. Write the question:
   One natural instruction at 00:00. Emphasizes unique/
   different. No expected count revealed.

4. Write responses — one per NEW unique target:
   - Describe what distinguishes from prior targets.
   - count: cumulative unique count (1, 2, 3,...).
   - trigger_time = first appearance (MM:SS).
   - Under 20 words. NEVER count re-appearances.

5. Classify: trigger_type, audio_dependency.

## Output (single JSON object, no markdown)
Fields: status, question, question_time, audio_dependency,
responses[] with: trigger_time, response, count,
trigger_type, trigger_type_reason, event_description.
If no dedup pattern or <3 targets: {"status":"skip",...}

## Rules
- Timestamps MM:SS in [00:00,{duration_mmss}].
- At least 3 responses. Each under 20 words.
- NEVER count re-appearances. Count increments by 1.
- Appear-disappear-reappear pattern REQUIRED.
- Targets spread across time. Video is ground truth.
- Skip freely — quality over quantity.
\end{verbatim}
\end{tcolorbox}

\begin{tcolorbox}[colback=gray!5, colframe=gray!80, title={\textbf{Sequential Step Instruction (SSI)}}, fonttitle=\small\color{white}, colbacktitle={rgb,255:red,64;green,64;blue,64}, boxrule=0.5pt, arc=2pt, breakable]
\small
\begin{verbatim}
## Task
"Sequential Step Instruction" — user states a learning
goal (following a tutorial). Assistant monitors in real
time and tells user what to do next at the right moment.
Tests: (1) understanding progress, (2) reasoning about
next step, (3) timely instructions. Distinction from EN:
EN is retrospective; SSI is prospective.

## Input
1. Original video (ground truth).
2. Timestamped dense caption (reference).
Duration: {duration_mmss} ({duration_sec:.0f}s).

## Constraints
- Real-time — no future knowledge. Instructions based on
  observations + domain knowledge.
- ONLY tutorials: cooking, DIY, repair, beauty, exercise.
  NOT: interviews, vlogs, news, reviews, sports.
  If not a replicable process, return skip.

## Steps

1. Determine suitability:
   Clear goal? Sequential steps? Observable?
   If any = NO, return skip.

2. Write the question:
   One natural instruction at 00:00. User wants to
   follow along. States learning goal. No spoilers.

3. Write responses — one per step transition:
   Timing: previous step completed, next not started.
   - Actionable instruction (WHAT + HOW).
   - Key parameters (quantities, temps, times).
   - Instructional language ("Now add...", "Next...")
     NOT descriptive ("He is adding...").
   - Verified by video. trigger_time (MM:SS).
   - Under 40 words. Chronological.

4. Classify: trigger_type, audio_dependency.

## Output (single JSON object, no markdown)
Fields: status, question, question_time, audio_dependency,
responses[] with: trigger_time, response, trigger_type,
trigger_type_reason, event_description.
If not tutorial: {"status":"skip","reason":"..."}

## Rules
- Timestamps MM:SS in [00:00,{duration_mmss}].
- At least 3 responses. Each under 40 words.
- Instructions MUST match what happens in video.
- Instructional language (address user as "you").
- Trigger at step transitions, not mid-step.
- Skip freely — most videos are NOT tutorials.
\end{verbatim}
\end{tcolorbox}

\section{Limitations, Broader Impacts, and Licenses}
\label{app:ethics}

\subsection{Limitations}
\label{app:limitations}

All questions and ground-truth annotations in \benchname are written in English, which limits its applicability for evaluating multilingual or non-English proactive streaming models. Extending the benchmark to additional languages is left for future work.

\subsection{Broader Impacts}
\label{app:impacts}

\paragraph{Positive impacts.}
\benchname advances research on proactive AI assistants by providing the first standardized evaluation covering omni-modal perception, proactive responding, and diverse video understanding tasks. It facilitates fair comparison across models and identifies concrete capability gaps, guiding future research directions.

\paragraph{Potential risks.}
As with any video understanding benchmark, improved model capabilities could in principle be applied to unintended contexts. However, our benchmark evaluates general-purpose understanding abilities and does not introduce domain-specific risks beyond those inherent to the underlying models.

\paragraph{Mitigation.}
We release the benchmark under a CC BY-NC 4.0 license, prohibiting commercial use. The dataset contains only publicly available YouTube videos from existing research datasets, with no personally identifiable information in annotations.

\subsection{Licenses}
\label{app:licenses}

\begin{itemize}[leftmargin=*,nosep]
  \item \textbf{LongVALE}~\cite{longvale}: CC-BY-NC-SA-4.0
  \item \textbf{COIN}~\cite{coin}: CC BY-NC 4.0
  \item \textbf{\benchname} (our benchmark): CC BY-NC 4.0
  \item \textbf{Evaluation code}: MIT License
\end{itemize}

Our license (CC BY-NC 4.0) is compatible with the source dataset licenses. All source datasets are properly cited and their terms of use are respected.


\end{document}